\documentclass[%
 aip,
 amsmath,amssymb,
 reprint,%
]{revtex4-1}

\usepackage{hyperref}
\usepackage{graphicx}
\usepackage{xcolor}
\usepackage{ulem}
\usepackage{siunitx}
\usepackage{dcolumn}
\usepackage{bm}

\usepackage[utf8]{inputenc}
\usepackage[T1]{fontenc}
\usepackage{mathptmx}
\usepackage{etoolbox}

\makeatletter
\def\@email#1#2{%
 \endgroup
 \patchcmd{\titleblock@produce}
  {\frontmatter@RRAPformat}
  {\frontmatter@RRAPformat{\produce@RRAP{*#1\href{mailto:#2}{#2}}}\frontmatter@RRAPformat}
  {}{}
}%
\makeatother
\begin{document}

\preprint{AIP/123-QED}

\title[]{RheOFormer: A transformer model for simulation of complex fluids and flows}
\author{Maedeh Saberi}
\affiliation{\mbox{Department of Mechanical and Industrial Engineering, Northeastern University, Boston, Massachusetts 02115, USA}}
\author{Amir Barati Farimani}
\affiliation{\mbox{Department of Mechanical Engineering, Carnegie Mellon University, Pittsburgh, Pennsylvania 15213, USA}}

\author{Safa Jamali}%
 \email{s.jamali@northeastern.edu}
\affiliation{\mbox{Department of Mechanical and Industrial Engineering, Northeastern University, Boston, Massachusetts 02115, USA}}

\date{\today}

\begin{abstract}
The ability to model soft materials under flow is key in designing processes and materials with targeted properties. This generally requires solution of the internal stress tensor, which is related to the deformation tensor through time-dependent nonlinear constitutive models. Traditional methods often suffer from prohibitive computational demands and poor scalability to other instances. Here, we introduce Rheological Operator Transformer (RheOFormer), an attention-based operator learning method to efficiently learn different spatial interactions and features of complex flows. We benchmark RheOFormer in a variety of complex fluids and complex domains against ground truth solutions. RheOFormer accurately learns both scalar and tensorial nonlinear mechanics of different complex fluids and predicts the spatio-temporal evolution of their flows, even when trained on limited datasets. 
By further quantifying both the data requirements and the failure boundaries of operator transformers for non-Newtonian flows, this work establishes the conditions under which attention-based surrogates are a reliable substitute for direct simulation. Its consistent performance without architectural modification across distinct constitutive models, geometries and flow protocols, together with the computational efficiency, establishes RheOFormer as a robust neural surrogate for accelerating predictive complex fluid simulations, advancing data-driven experimentation, and enabling real-time process optimization across a wide range of applications.
\end{abstract}

\maketitle

\section{Introduction}
Soft materials and complex fluids are ubiquitous in nature\cite{jerolmack2019viewing}, biology\cite{levental2007soft}, food\cite{mezzenga2005understanding}, additive manufacturing\cite{truby2016printing}, and many other applications. Mechanics of complex fluids such as polymeric and particulate systems include nonlinear rate- and time-dependent response to an applied deformation that manifests in viscoelasticity, viscoplasticity, and/or thixotropic effects\cite{bird1987dynamics, macosko1994rheology}. Hence, the ability to model and simulate these non-Newtonian fluid dynamics under various flow conditions is pivotal to advancing numerous scientific disciplines. Despite extensive theoretical and computational advancements, modeling the full rheological response of complex fluids remains challenging. Traditional numerical approaches, such as finite-element or finite-volume\cite{alves2021numerical}, are well-known methods of solving differential equations by discretizing the solution domain and converting the respective constitutive relations into finite-dimensional problems. These methods are often computationally intensive, particularly when addressing high-dimensional, history-dependent problems\cite{walters2003distinctive}. They are also constrained by specific boundary and initial conditions, necessitating a full re-computation for each new scenario. These challenges are further amplified when considering real-world flow protocols and geometries that can induce large stress gradients and complex time-evolving structures\cite{keunings1990progress}.

In recent years, data-driven techniques have increasingly been leveraged to address the complexities inherent in modeling fluid dynamics\cite{brunton2020machine,taira2017modal, cai2021physics, lino2023current}. One promising direction has been the integration of physical laws into neural network frameworks, leading to the development of Physics-Informed Neural Networks (PINNs)\cite{raissi2019physics,karniadakis2021physics}. These architectures enforce the governing partial differential equations directly within the learning process, thereby reducing the strict need for large datasets\cite{mahmoudabadbozchelou2022nn}. Expanding upon this idea, frameworks such as Rheology-Informed Neural Networks (RhINNs) have been proposed, specifically tailoring the learning process to solve rheologically-relevant constitutive equations\cite{mahmoudabadbozchelou2021rheology,lennon2023scientific, thakur2024viscoelasticnet, thakur2024inverse, mahmoudabadbozchelou2021data, dabiri2025detailed}. These models have demonstrated significant success in both forward simulations and inverse problems\cite{raissi2020hidden}, enabling the identification of complex rheological parameters from limited experimental data\cite{mahmoudabadbozchelou2024unbiased, sato2025rheo}. Nonetheless, despite their strength in implementation and accuracy, PINNs often require re-optimization when applied to different instances of a given constitutive equation, such as changes in material parameters or boundary conditions, limiting their scalability across diverse problem settings\cite{krishnapriyan2021characterizing}. Parallel to the development of physics-informed approaches, purely data-driven partial differential equation (PDE) solvers have emerged by learning solutions directly from observational data, without requiring explicit knowledge of the underlying governing equations\cite{khoo2021solving}. These approaches typically utilize supervised deep learning architectures, incorporating inductive biases appropriate to the structure of the problem. For example, convolutional layers are employed for structured grids\cite{zhu2018bayesian, bhatnagar2019prediction, kochkov2021machine}, while graph neural networks capture unstructured local relationships within complex systems\cite{belbute2020combining, ogoke2021graph, aminimajd2025scalability}. These data-driven methods have found increasing application in rheology, enabling faster material characterization and accelerating numerical simulations. Such frameworks offer an attractive pathway towards predictive modeling, particularly in situations where explicit constitutive models are either unknown or difficult to derive\cite{mangal2024data}. Despite their remarkable promise, traditional deep learning methods often suffer from key limitations, notably their restriction on the input resolution and geometry of the training data, which necessitates retraining when encountering new scenarios. These challenges motivate the exploration of more flexible and physics-compatible approaches, such as Neural Operators, which aim to generalize across families of problems without requiring retraining\cite{azizzadenesheli2024neural,kovachki2023neural}.

To address the limitations of instance-specific models, Neural Operators have emerged as powerful algorithms in learning mappings between entire function spaces, rather than discrete points. Neural Operators such as the Fourier Neural Operator (FNO)\cite{li2020fourier} and DeepONet\cite{lu2021learning} provide a framework capable of learning the solution operators of complex PDEs using a practical realization of the general universal nonlinear operator approximation theorem\cite{chen1995universal}. These advances in operator learning have sparked significant research interest, largely due to their ability to generalize across a class of PDEs without requiring retraining when faced with new boundary or initial conditions\cite{wen2022u, cai2021deepm, li2020neural, li2020multipole}. In the context of rheology, Neural Operators have demonstrated exceptional capabilities in learning families of constitutive models across varying flow protocols and fluid types\cite{mangal2025learning}. Compared to conventional neural networks, Neural Operators offer enhanced generalization, flexibility across different geometries, and computational efficiency in real-time predictions. Nevertheless, challenges remain in scaling these architectures for highly nonlinear, history-dependent behaviors typical of complex fluids, and ensuring that physical constraints are consistently honored during learning.

Complementing these advancements is the rapid rise of generative models—such as autoencoders\cite{pinaya2020autoencoders,zhou2024masked}, attention-based transformers\cite{vaswani2017attention} and diffusion models\cite{yang2023diffusion}. Growing interest in attention-based architectures, initially popularized by breakthroughs in natural language processing\cite{chorowski2015attention}, has led to their adoption across different domains\cite{velickovic2017graph,rodriguez2022physics, boya2024physics}. Two main pathways to solving PDEs via attention-based architectures have been developed: using attention to encode spatial structures and patterns\cite{kissas2022learning, shao2022sit, hao2023gnot}, and employing it for modeling temporal dynamics\cite{geneva2022transformers, han2022predicting}. Learning the temporal evolution in spatio-temporal PDE systems remains a significant challenge due to its high memory requirements and computational overhead. To alleviate this burden, latent time-marching strategies have been introduced by encoding system dynamics into a lower-dimensional latent representation. Then, time evolution can be efficiently learned using linear propagators based on Koopman operator theory\cite{schmid2022dynamic, mezic2013analysis, lusch2018deep, morton2018deep}.

In this work, we introduce RheOFormer, an attention-based transformer model for the solution of rheologically relevant PDEs, leveraging the architecture of OFormer\cite{li2022transformer} and operator learning. 
The choice of a transformer-based architecture for modeling complex fluid flows is motivated by several fundamental advantages. Transformers, through their attention mechanisms, can capture long-range spatial dependencies without the locality constraints of convolutional or graph-based architectures, which is critical for flows where upstream conditions influence downstream behavior across extended domains. Additionally, the flexibility of attention mechanisms in handling variable-length inputs and arbitrary query locations makes transformers naturally suited to problems involving irregular geometries and multi-scale physics, both hallmarks of non-Newtonian fluid dynamics. Among existing approaches for learning PDE solution operators, Fourier Neural Operators (FNO)\cite{li2020fourier}, while powerful for learning operator mappings, only operate on fixed regular grids and lack flexibility when applied to irregular geometries such as contractions or flows around obstacles.
Graph neural network-based methods\cite{sanchez2020learning, pfaff2020learning} offer geometric adaptability but do not incorporate efficient temporal propagation mechanisms, requiring costly fully unrolled training over long time horizons. Within the family of transformer-based PDE solvers, standard attention architectures\cite{kissas2022learning, hao2023gnot} typically couple input and output to the same discretization grid, limiting their applicability when solution queries are needed at arbitrary spatial locations.

OFormer was selected as the backbone of RheOFormer because it uniquely addresses all of these challenges. Its cross-attention mechanism decouples input and output discretizations, and its softmax-free attention variants connect directly to integral operator approximation. Also, by leveraging its latent time-marching mechanism, OFormer allows efficient propagation of long-horizon temporal dynamics in latent space without prohibitive memory costs while capturing spatial patterns using the attention structure. These properties make this architecture particularly well-suited for the history-dependent, spatially complex problems characteristic of non-Newtonian fluid dynamics.

We systematically examine RheOFormer's ability to learn diverse rheological behaviors by testing it on a broad spectrum of problems, ranging from simple ordinary differential equations to complex coupled PDEs in different domains. Benchmarking against numerical solutions, we aim to highlight RheOFormer’s flexibility and also the practical challenges and considerations in applying operator transformers to history-dependent non-Newtonian flows.

\section{RheOFormer Architecture}\label{RheOFormer}

Built upon the original OFormer\cite{li2022transformer} algorithm, RheOFormer employs an encoder-decoder architecture reminiscent of the original Transformer introduced by Vaswani et al.\cite{vaswani2017attention}. 
At a high level, the architecture operates in three stages. 
The \textit{encoder} takes spatially distributed physical field data (e.g., velocity and stress) and learns how different spatial locations influence one another, producing a compressed representation of the input state. 
The \textit{decoder} takes the coordinates of desired output locations and retrieves relevant information from the encoded representation to reconstruct field values at those points. 
Finally, the \textit{propagator} advances the decoded latent state forward in time through a sequence of lightweight feed-forward steps, enabling multi-step temporal predictions without re-encoding the input at each time step. The mathematical details of each component are described below.

\begin{figure*}
\centering
\includegraphics[width=\textwidth]{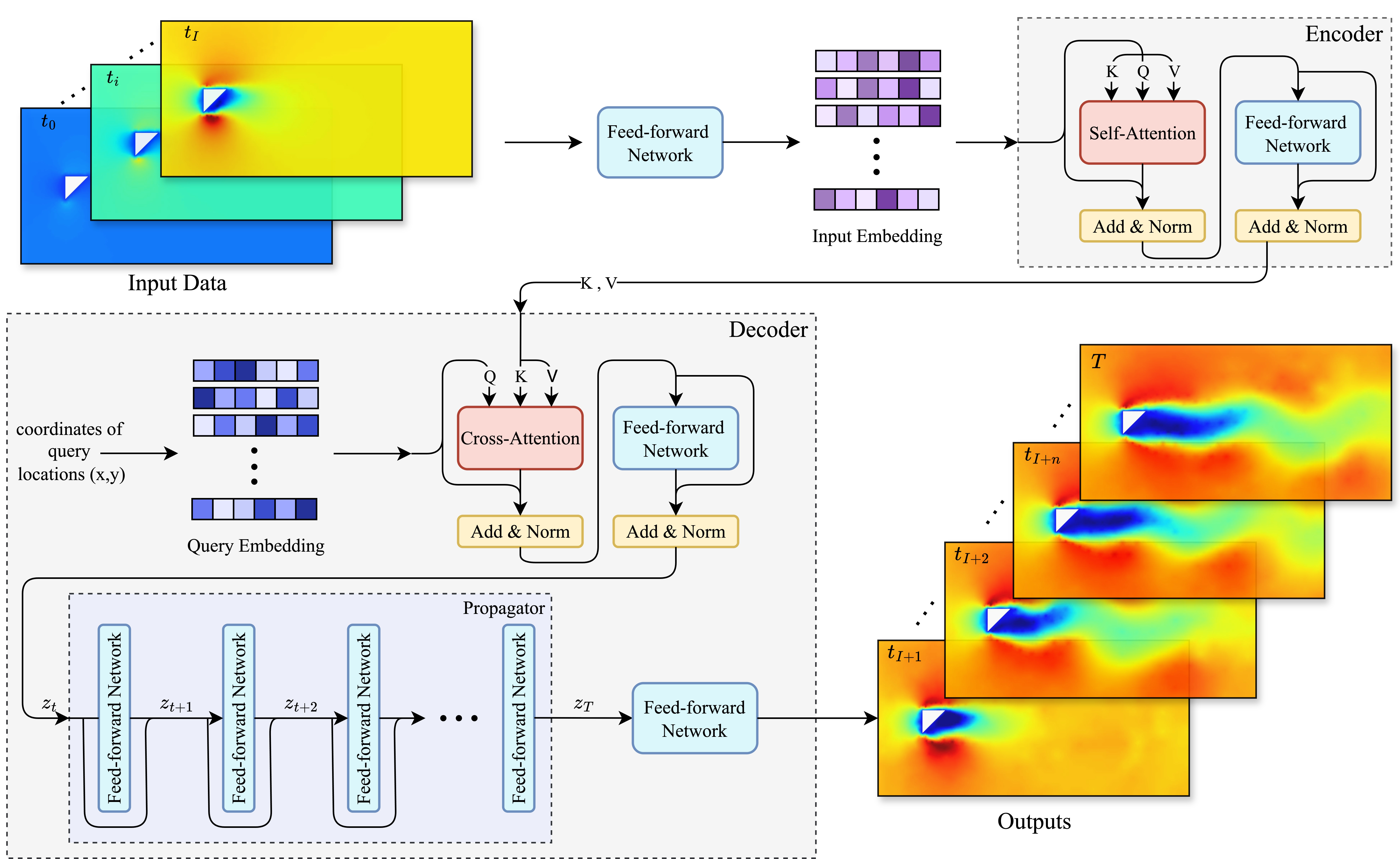}
\caption{Schematic representation of the RheOFormer architecture for operator learning in complex fluid flows. The model takes spatially distributed input fields (e.g., velocity, stress) and processes them through a feed-forward network and self-attention encoder to capture spatial dependencies. Query locations are encoded and passed through a decoder where cross-attention mechanisms integrate information from the encoded inputs and output points. A latent time-marching propagator then recursively evolves the system through time in latent space, and the final latent representation is decoded to produce the predicted output field (e.g., velocity or stress). An example prediction is shown for flow past a triangular obstacle.}
\label{fig:method}
\end{figure*}

\textbf{Encoder}. The encoder module consists of three main subcomponents (Figure \ref{fig:method}). Initially, an input encoder integrates the sampled values of the input function $a(x_i)$ along with their respective coordinates $\{x_i\}^n_{i=1}$ as input features. 
The input to the encoder is a set of spatially distributed field observations $a(x_i, t_k)$ sampled at $n$ spatial locations $\{x_i\}_{i=1}^{n}$ over the first $I$ time snapshots, where $a$ is a vector-valued function containing the physical variables of interest (e.g., velocity field). The specific inputs and outputs for each case are detailed in the \hyperref[results]{Results} section.

These sampled values, concatenated with their spatial coordinates, form the input features. The input features are then transformed into high-dimensional embeddings via a feed-forward network.
Subsequently, these embeddings are passed through a self-attention module, which processes the embeddings by generating query ($Q$), key ($K$), and value ($V$) representations. After the self-attention operation, the outputs undergo an ``Add \& Norm" step that adds residual connections and applies layer normalization. This is followed by another feed-forward network to further refine the representations, and a final ``Add \& Norm" step that again incorporates residual connections and normalization to stabilize learning.

Self-attention, also known as scaled dot-product attention, is a mechanism enabling the model to weigh the importance of different input elements dynamically. Unlike traditional attention mechanisms that focus on fixed positional relevance, self-attention allows each position in a sequence to attend to all other positions, facilitating the capture of intricate, context-dependent interactions and correlations. This flexibility significantly enhances the model's capability in understanding complex patterns and dependencies inherent in sequential or spatial data.

Standard attention mechanisms\cite{vaswani2017attention, bahdanau2014neural, graves2014neural, luong2015effective} operate upon three sets of vectors, termed queries $(Q)$, keys $(K)$, and values $(V)$. Cao\cite{cao2021choose} introduced an interpretation wherein each column of query/key/value matrices corresponds to evaluations of learned basis functions at discrete points.
For instance, elements such as $V_{ij}$ represent evaluations of the $j$-th basis function at grid point $x_i$, i.e., $V_{ij}=v_j(x_i)$, similarly for $Q$ and $K$. Throughout, bold symbols with subscripts denote row vectors at a grid point (e.g., $\mathbf{q}_i$ is the row of $Q$ at points $x_i$), while bold symbols with superscripts denote column vectors of a basis function (e.g., $\mathbf{v}^j$ is the j-th column of V), and parenthetical subscripts extract scalar entries: $(\mathbf{v}^j)_s = v_j(x_s)$. Leveraging this basis-function perspective, Cao\cite{cao2021choose} proposed softmax-free attention variants that approximate integral operators via numerical quadrature, where dot products ($\mathbf{q}_i\cdot \mathbf{k}_s$) approximate a learnable kernel $\kappa(x_i,x_s)$.

\begin{equation}
\begin{aligned}
\text{Fourier type:} \quad 
(\mathbf{z}_i)_j = \frac{1}{n} \sum_{s=1}^{n} (\mathbf{q}_i \cdot \mathbf{k}_s)(\mathbf{v}^j)_s \\
\approx \int_\Omega \kappa(x_i, \xi) v_j(\xi) \, d\xi
\end{aligned}
\end{equation}

Analogously, $(\mathbf{q}^l)_i = q_l(x_i)$ and the dot product $(\mathbf{k}^l \cdot \mathbf{v}^j) = \sum_{s=1}^{n} k_l(x_s)\,v_j(x_s)$ denotes the discrete inner product over all $n$ grid points in the Galerkin form.

\begin{equation}
\begin{aligned}
\text{Galerkin type:} \quad 
(\mathbf{z}^j)_i = \sum_{l=1}^{d} \frac{(\mathbf{k}^l \cdot \mathbf{v}^j)}{n} (\mathbf{q}^l)_i \\
\approx \sum_{l=1}^{d} \left( \int_\Omega k_l(\xi) v_j(\xi) \, d\xi \right) q_l(x_i)
\end{aligned}
\end{equation}

These integral-based attention mechanisms serve as efficient and powerful building blocks for PDE operator learning, enabling simplified computation as $Z=Q(K^T V)/n$ due to the associative nature of matrix multiplication.

\textbf{Decoder.} The decoder module takes the coordinates of query locations $\{y_j\}_{j=1}^{m}$ (the points at which the output solution is to be evaluated) and transforms them into latent embeddings that are subsequently used via cross-attention. 
Since this operation acts on raw spatial coordinates, it is susceptible to spectral bias, whereby coordinate-based neural networks preferentially learn low-frequency functions and struggle to represent fine-scale spatial features\cite{tancik2020fourier, wang2021eigenvector}.

To address this, a random Fourier projection\cite{rahimi2007random} is applied to map each coordinate into a high-dimensional space of sinusoidal features spanning a broad range of frequencies. This enables the network to capture both smooth large-scale structures and  fine-scale features  that commonly arise in physical fields, such as stress concentrations near boundaries or velocity variations across shear layers.
This projection is defined as:
\begin{equation}
\gamma(Y) = [\cos(2\pi Y B),\; \sin(2\pi Y B)]
\end{equation}
where $Y = [y_1, y_2, \dots, y_m]^T \in \mathbb{R}^{m \times d_1}$ is the matrix of query coordinates, with each $y_j \in \mathbb{R}^{d_1}$ representing the Cartesian coordinates of the $j$-th query point ($d_1$ being the spatial dimension, e.g., $d_1=2$ for two-dimensional problems). The matrix $B \in \mathbb{R}^{d_1\times d_2}$, with entries drawn independently from $\mathcal{N}(0,\sigma^2)$, projects each $d_1$-dimensional coordinate into a $d_2$-dimensional Fourier feature space, yielding an embedded representation of $\gamma(Y) \in \mathbb{R}^{m \times 2d_2}$.
The Fourier features are then passed through subsequent feed-forward layers to produce the query embeddings.

The embedded query coordinates are then passed to a cross-attention module that integrates system-level information from the encoder. While self-attention in the encoder operates on a single set of embedded features, restricting its input and output to the same discretization grid $\{x_i\}_{i=1}^{n}$, cross-attention decouples the two: the query matrix $Q$ is constructed from the Fourier-embedded query locations $\{y_j\}_{j=1}^{m}$, while the key and value matrices $K$ and $V$ are derived from the encoder output defined on $\{x_i\}_{i=1}^{n}$. This allows the decoder to evaluate the solution at arbitrary query points independent of the input grid. Using the learned-basis interpretation introduced above, the cross-attention output takes the form:
\begin{equation}
z_s(y_j) = \sum_{l=1}^{d}\left(\frac{1}{n}\sum_{i=1}^{n}k_l(x_i)\,v_s(x_i)\right)q_l(y_j)
\end{equation}
where $k_l(\cdot)$ and $v_s(\cdot)$ are basis functions defined on the input discretization $\{x_i\}_{i=1}^{n}$, and $q_l(\cdot)$ on the query discretization $\{y_j\}_{j=1}^{m}$.

\textbf{Propagator}. To model time-dependent problems, a latent-space propagator introduced by Li et al.\cite{li2022transformer} is employed in this work. While direct augmentation of input grids with explicit temporal coordinates\cite{raissi2019physics} can lead to suboptimal performance and require excessive parameterization\cite{krishnapriyan2021characterizing}, autoregressive encoder-processor-decoder (EPD) schemes\cite{brandstetter2022message, stachenfeld2021learned} offer more effective training. Despite their effectiveness, fully unrolled training approaches carry prohibitive memory costs of order $O(tn)$, with \textit{t} the length of the time horizon and \textit{n} the number of model parameters. To overcome these challenges, a recurrent, sequence-to-sequence\cite{sutskever2014sequence} latent-space propagation strategy is adopted in this study (illustrated in Figure \ref{fig:method}). In contrast to conventional methods, the dynamics are propagated entirely in the latent space, significantly reducing memory usage since the encoder operates only once. Given the initial latent encoding $z_0$ obtained from input embeddings via cross-attention, the latent state is recursively advanced through a residual propagator $N(\cdot)$, formulated as $z_{t+1} = z_t + N(z_t)$. 

Although several architectures could parametrize the propagator, a simple point-wise feed-forward network shared across query locations and time steps is sufficient in practice, indicating that the original PDE can be effectively approximated by a fixed-interval ODE in latent space. Ultimately, another feed-forward network is utilized to decode the propagated latent state $z_t(x)$ into the predicted observable function values $u(x,t)$.

\textbf{Loss Function}. 
To train RheOFormer, we employ a relative $L^2$ loss that measures the discrepancy between predicted and ground truth fields, normalized by the magnitude of the ground truth itself. This normalization is critical in the context of complex fluid simulations, where the physical fields of interest (velocities and stress components) can span vastly different magnitudes across channels, spatial locations, and time steps. An absolute loss metric would disproportionately weight channels or regions with inherently larger magnitudes, effectively ignoring the accuracy of smaller-valued but physically important quantities such as off-diagonal stress components. By normalizing the error with respect to the ground truth norm, the relative loss ensures that each physical field contributes proportionally to the training signal, regardless of its absolute scale. For instance, in viscoelastic flow problems, normal stress components can be orders of magnitude smaller than the streamwise velocity, yet they carry essential information about the elastic response of the fluid. The relative loss naturally balances these disparate scales without requiring manual tuning of channel weights.

Formally, let $\hat{\mathbf{f}} \in \mathbb{R}^{B \times T \times N \times C}$ and $\mathbf{f} \in \mathbb{R}^{B \times T \times N \times C}$ denote the predicted and ground truth solution tensors, respectively, where $B$ is the batch size, $T$ the number of temporal frames, $N$ the number of spatial nodes, and $C$ the number of output channels (e.g., $u_x, u_y, \sigma_{xx}, \sigma_{yy}, \sigma_{xy}$). The relative loss is defined as:

\begin{equation}
\mathcal{L} = \frac{1}{B \cdot C}  \sum_{b=1}^{B} \sum_{c=1}^{C} \frac{\| \hat{\mathbf{f}}_{b,:,:,c} - \mathbf{f}_{b,:,:,c} \|_2}{\| \mathbf{f}_{b,:,:,c} \|_2 + \epsilon}
\end{equation}

where the $L^2$ norm is computed jointly over the temporal and spatial dimensions:

\begin{equation}
\| \mathbf{f}_{b,:,:,c} \|_2 = \sqrt{\sum_{t=1}^{T} \sum_{n=1}^{N} \mathbf{f}_{b,t,n,c}^2}
\end{equation}

for any field $\mathbf{f}$, and $\epsilon$ is a small constant introduced to prevent numerical instability in regions where the ground truth approaches zero. By computing the norm over both time and space simultaneously, the loss captures the integrated spatio-temporal prediction error for each channel, thereby encouraging the model to maintain accuracy across the entire simulation trajectory rather than at isolated snapshots. The summation over channels ensures that all physical fields are jointly optimized during training.

It is worth noting that the choice of relative loss formulation requires careful consideration when applied to physical problems. The original OFormer\cite{li2022transformer} employed a pointwise relative loss, in which the error at each individual grid point $(b,t,n,c)$ is normalized by the local ground truth value at that same point:

\begin{equation}
\mathcal{L}_{\text{pointwise}} = \frac{1}{B \cdot T \cdot N \cdot C} \sum_{b,t,n,c}
\frac{(\hat{\mathbf{f}}_{b,t,n,c} - \mathbf{f}_{b,t,n,c})^2}{\mathbf{f}_{b,t,n,c}^2 + \epsilon}
\end{equation}

For the benchmark problems considered in that work, this formulation performed well. However, in complex fluid simulations, pointwise normalization introduces a fundamental ill-conditioning: wherever the ground truth field approaches zero (e.g., at stagnation points, or near the walls), the local denominator $\mathbf{f}_{b,t,n,c}^2$ becomes vanishingly small, causing the pointwise relative error to diverge even when the absolute prediction error is negligibly small. These near-zero regions then dominate the gradient signal during training, effectively directing the optimizer to focus on physically unimportant areas of the domain at the expense of regions where accuracy is most critical. This pathology persists even when data normalization is applied, as many physical fields inherently contain regions of near-zero magnitude that cannot be eliminated through pre-processing.

The global norm formulation adopted here circumvents this issue entirely. Because the normalization is performed over the aggregate $L^2$ norm of the entire spatio-temporal field, the denominator is dominated by regions with substantial signal magnitude. Near-zero points contribute negligibly to both the numerator and denominator, and therefore do not distort the loss landscape or the resulting gradient signal. In essence, the pointwise formulation asks ``what is the relative error at each individual point?". This is an ill-posed question where the field vanishes; while the global norm formulation asks ``what fraction of the total field is captured by the prediction?", a well-conditioned question regardless of local field values. This distinction proved essential for achieving stable convergence and high prediction accuracy across the diverse set of physical problems considered in this work, where output fields span multiple orders of magnitude and routinely contain spatially extended near-zero regions.

\section{Results and Discussion}\label{results}

The benchmarking problems are chosen to span increasing levels of complexity: from scalar to tensorial coupled ODEs, and finally to full spatio-temporal PDE systems in two-dimensional flow geometries. 
This progression systematically evaluates RheOFormer across fundamentally different problem structures, thereby demonstrating its applicability beyond any single class of governing equations. 

Accordingly, this study is structured into two sets of experiments: (1) ``rheometric flows,'' in which a given kinematic is applied through input deformation rates corresponding to classical viscometric flows and the resulting stress response is modeled, and (2) ``canonical flows,'' where the temporal evolution of the entire velocity and stress fields is predicted in benchmark flow geometries. 
Specifically, the 4:1 contraction geometry is a classical test case in computational rheology\cite{alves2021numerical, keunings1990progress, lennon2023scientific}, known for producing large stress gradients near the corners, while flow past an obstacle presents additional challenges by introducing unsteady wake dynamics and shear layer formation.

\begin{figure}
\centering
\includegraphics[width=8.7cm]{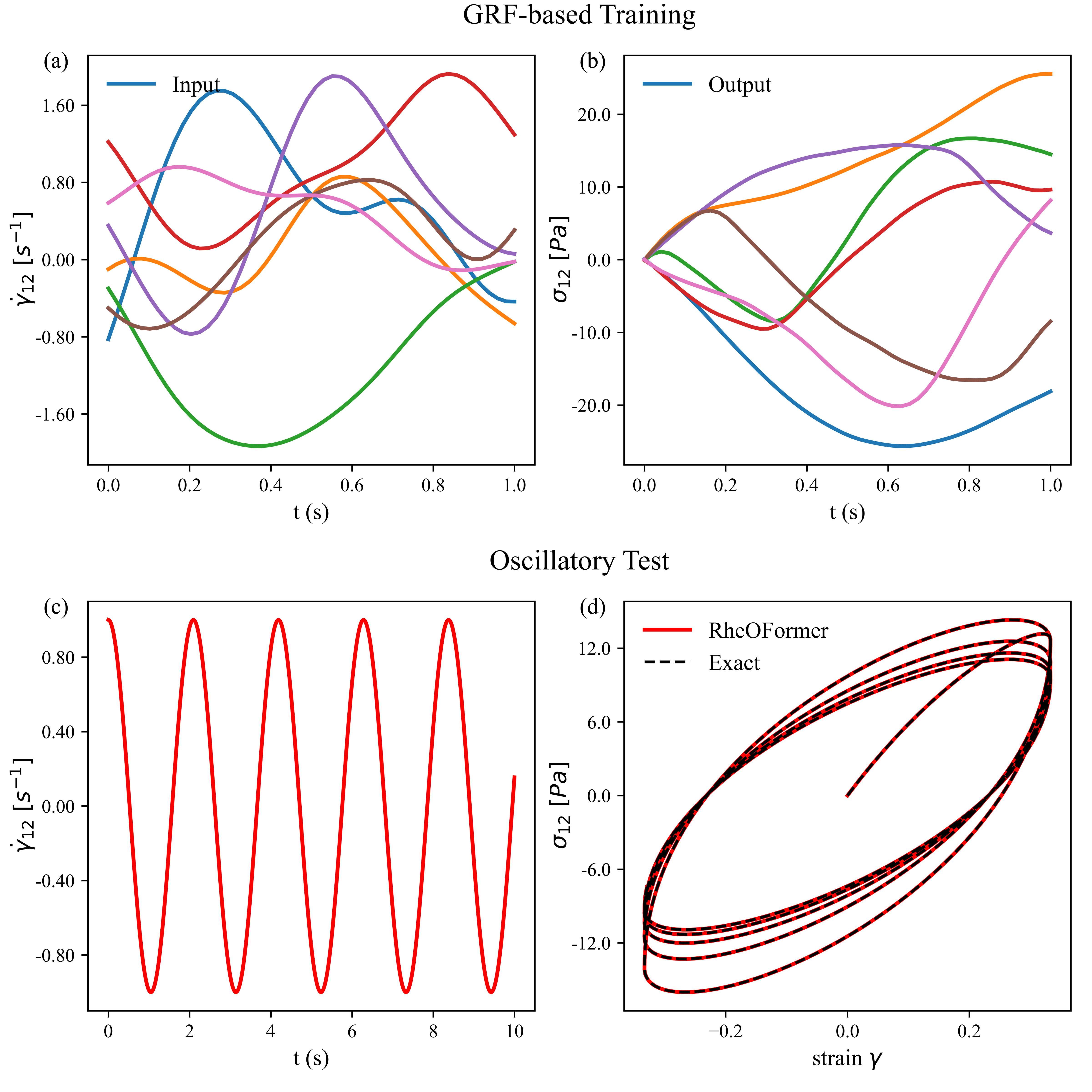}
\caption{RheOFormer training and predictions for the TEVP constitutive model.
(a) Sample shear rate profiles \( \dot{\gamma}_{12}(t) \) drawn from a Gaussian Random Field (GRF) used as training inputs. 
(b) Corresponding predicted shear stress response \( \sigma_{12}(t) \) (solid lines) compared with exact solutions (dashed) for each input case. 
(c) Applied oscillatory shear rate profile \( \dot{\gamma}_{12}(t) \) over extended time. 
(d) Predicted shear stress \( \sigma_{12} \) versus strain \( \gamma \) under oscillatory shear (solid red), compared with ground truth solution (dashed black).
}
\label{fig:tevp}
\end{figure}

\begin{figure}
\centering
\includegraphics[width=8.7cm]{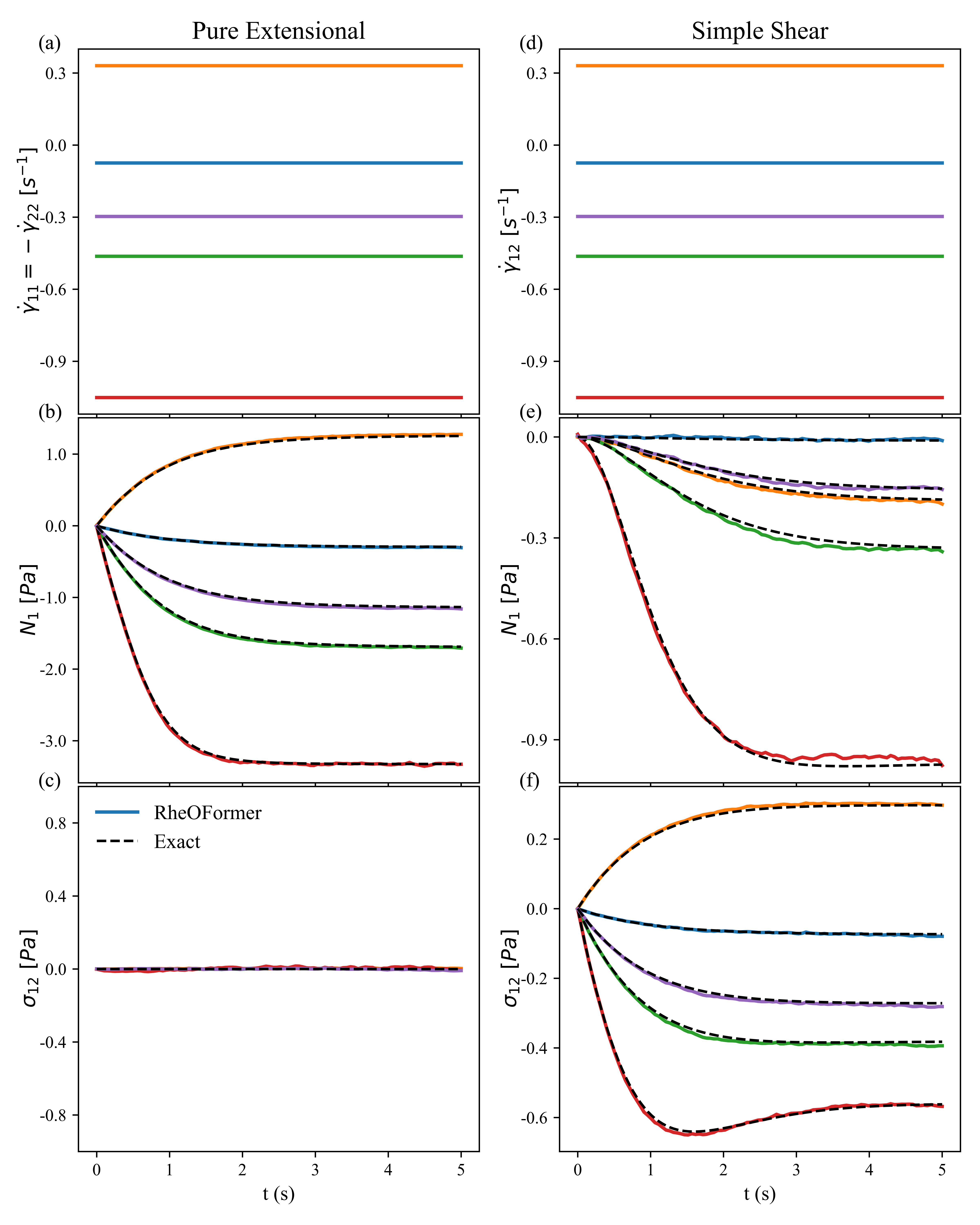}
\caption{RheOFormer predictions for the tensorial stress response of the Giesekus model under pure extensional (a–c) and simple shear flows (d–f).
The model was trained on $N_\mathrm{train}=7000$ samples in which all four velocity gradient components were drawn independently from Gaussian fields, and is tested on protocols with symmetry constraints that were absent from training.
Panels (a, d) show the applied constant deformation rate inputs. Panels (b, e) display the evolution of the first normal stress difference $N_1 = \sigma_{11} - \sigma_{22}$ over time, while (c, f) show the corresponding shear stress component $\sigma_{12}(t)$. Solid lines show RheOFormer predictions and dashed lines show the ground truth values.}
\label{fig:Giesekus}
\end{figure}

\subsection{Rheometric Flows}
In this section, RheOFormer is used to learn operators corresponding to ordinary differential equations (ODEs), specifically addressing complex cases involving coupled and tensorial ODE systems. Namely, a thixotropic elasto-visco-plastic fluid (TEVP) is modeled for the case of coupled ODEs, and a Giesekus fluid is modeled for the case of tensorial ODEs.
In both cases, training inputs were generated from Gaussian random fields (GRF) with a squared exponential kernel, $k(t,t') = \mathcal{S}^2\exp\!\left(-(t-t')^2/2\ell^2\right)$, with correlation time $\ell = 0.2~s$ and output scale $\mathcal{S} = 1.0$, where the output scale $\mathcal{S}$ controls the standard deviation of the GRF and thus the characteristic magnitude of the generated deformation rate profiles. 

GRF realizations were chosen as training inputs because they naturally span a broad range of temporal frequencies and amplitudes, exposing the model to a rich variety of transient dynamics without biasing it toward any particular flow protocol. Although these random input functions do not correspond to standard rheological tests, this diversity is exactly what enables generalization in functional form: the model learns to map arbitrary input shapes to their corresponding stress responses, rather than memorizing specific flow protocols. The effectiveness of this choice is supported by the test results presented below, where RheOFormer accurately predicts responses under both oscillatory shear and constant-rate flows, functional forms that are structurally distinct from the stochastic GRF profiles used during training.

For TEVP, the model was trained on 1000 GRF-based random realizations of input functions $\dot{\boldsymbol{\gamma}}_{12}(t)$.
Each training sample was resolved over the time interval $t \in [0, 10]$~s with 500 temporal points ($\Delta t = 0.02$~s). The GRF realizations produced shear rate profiles with amplitudes typically ranging from approximately $-3$ to $3$~s$^{-1}$. The model was trained to predict both the shear stress $\sigma_{12}(t)$ and the structure parameter $\lambda(t)$ simultaneously, thereby learning the full coupled dynamics of the TEVP system. The TEVP model parameters used for this case are: elastic modulus $G = 40$~\si{\pascal}, yield stress $\sigma_y = 10$~\si{\pascal}, solvent viscosity $\eta_s = 10$~\si{\pascal}$\cdot$\si{\second}, plastic viscosity $\eta_p = 5$~\si{\pascal}$\cdot$\si{\second}, structural buildup rate $k_+ = 0.1$~\si{\per\second}, and structural breakdown rate $k_- = 0.3$~\si{\per\second} (see \hyperref[TEVP]{TEVP Model} for formulation and parameter definitions).

Figure \ref{fig:tevp}(a,b) shows representative sets of GRF-generated shear rate input profiles $\dot{\gamma}_{12}(t)$, and their corresponding shear stress output profiles $\sigma_{12}(t)$ used for training purposes. Having trained on similar series of GRF-generated input/output functions, the RheOFormer was then tested on rheometric flows whose functional forms were not observed during the training.
Figure \ref{fig:tevp}(c,d) shows the RheOFormer performance in predicting the TEVP outputs for a representative oscillatory shear test with $\dot{\gamma}_{12}(t) = \cos(3t)$ (amplitude $1.0$~s$^{-1}$, angular frequency $\omega = 3$~rad/s). The amplitude of this test signal falls within the range covered by the training data; however, the smooth periodic functional form is fundamentally different from the stochastic GRF profiles used during training, testing the model's ability to generalize across input function classes.
The predicted outputs (red solid lines) are compared against ground truth solutions of a TEVP constitutive equation (black dashed-lines). Specifically, Figure \ref{fig:tevp}(c) represents the applied shear rate profile $\dot{\gamma}_{12}(t)$, and Figure \ref{fig:tevp}(d) is the shear stress response ($\sigma_{12}$) as a function of the applied strain ${\gamma}_{12}$. Overall, predictions obtained from the RheOFormer closely match the exact solutions, demonstrating accurate predictions within the tested parameter regime.

\begin{figure*}[t]
\centering
\includegraphics[width=\linewidth]{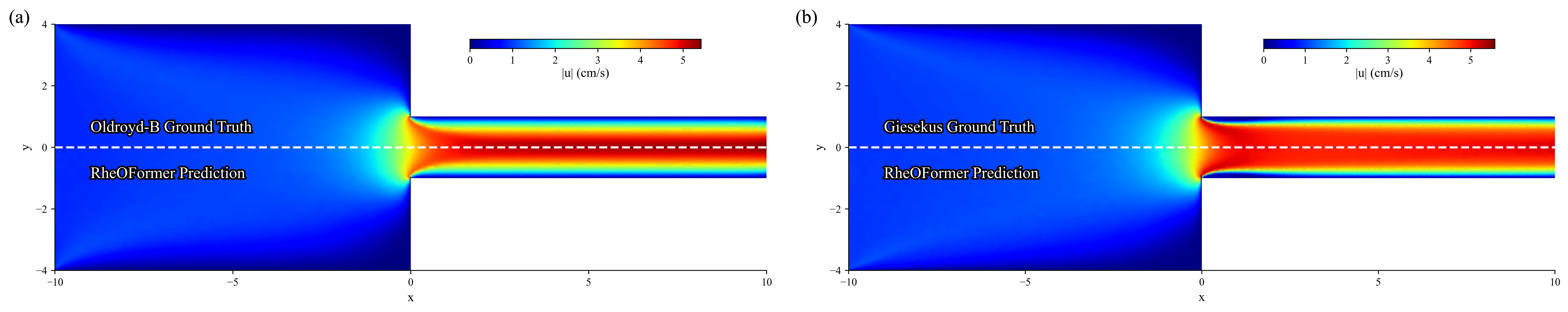}
\caption{Comparison of RheOFormer predictions and ground truth solutions for viscoelastic flows through a 4:1 planar contraction channel. (a) Velocity magnitude field $|\mathbf{u}|$ for the Oldroyd-B fluid at the final time with $Wi=0.35$ and $Re=35$; (b) Corresponding results for the Giesekus fluid with $Wi=3.9$ and $Re=39$. In each figure, the upper half displays the numerically simulated ground truth solutions and the lower half shows the RheOFormer predictions. The cases shown correspond to $Wi$ values not included in the training set.}
\label{fig:channel}
\end{figure*}

RheOFormer is next employed to predict the rheological responses of a Giesekus fluid in its tensorial form. Unlike the scalar-input scenario, the Giesekus model features multiple input and output variables; specifically, the inputs consist of the four velocity gradient components $\nabla u_{11}(t)$, $\nabla u_{22}(t)$, $\nabla u_{12}(t)$, and $\nabla u_{21}(t)$, from which the deformation rate tensor is constructed (see \hyperref[Giesekus]{Giesekus Model} for formulation details). The outputs are the stress tensor components $\sigma_{11}$, $\sigma_{22}$, and $\sigma_{12}$.
For this case, 7000 training samples were generated, each with all four velocity gradient components drawn independently from GRF realizations using the same kernel parameters as the TEVP case (RBF kernel, length scale $l = 0.2$, output scale $\mathcal{S} = 1.0$). The GRF realizations produced velocity gradient amplitudes typically ranging from approximately $-3$ to $3$~s$^{-1}$, corresponding to Weissenberg numbers up to $Wi = \dot{\gamma}\tau_1 \approx 3$. Each sample was resolved over the time interval $t \in [0, 5]$~s with 100 temporal points ($\Delta t = 0.05$~s).
For the rheometric case, a Giesekus model without retardation ($\tau_2 = 0$) was used, with relaxation time $\tau_1 = 1$~s, elastic modulus $G_0 = \eta_0/\tau_1 = 1$~Pa, and mobility factor $\alpha = 0.8$ (see \hyperref[Giesekus]{Giesekus Model} for formulation details and parameter definitions).

Figure ~\ref{fig:Giesekus} demonstrates RheOFormer's predictions of the first normal stress difference $N_1 = \sigma_{11} - \sigma_{22}$, and the shear stress response ($\sigma_{12}$) for two canonical rheological tests: planar extensional flow and simple shear flow. 
For planar extensional flow, the velocity gradient tensor is diagonal, indicating elongation in one direction and equal contraction perpendicular to it ($\nabla u_{12} = \nabla u_{21} = 0, \nabla u_{11} = -\nabla u_{22} \neq 0$). In a simple shear flow, the velocity gradient tensor involves linear velocity variations in one direction, characterized by a single nonzero off-diagonal component ($\nabla u_{11} = \nabla u_{22} = \nabla u_{21} = 0$). In both test cases, constant deformation rates are applied, with magnitudes in the range of approximately $0.1$ to $0.9$~s$^{-1}$ (as shown in Figure ~\ref{fig:Giesekus}a,d), which falls within the amplitude range covered by the training data. 
As with the TEVP case, the generalization demonstrated here is primarily in functional form. The model trained on stochastic, simultaneously varying GRF inputs, accurately predicts the response under structured canonical flow conditions with constant rates and specific symmetry constraints that were never seen during training. The predictions (red curves) are compared against ground truth solutions (dashed lines) of a Giesekus fluid.

\subsection{Canonical Flows}
Having established RheOFormer's ability in learning and modeling complex fluids' response to rheometric flows, in the next step, viscoelastic flows are modeled in canonical and arbitrary geometries. This entails learning and predicting the full spatio-temporal dynamics of viscoelastic fluid flows under various physical settings and constitutive relations. 
We first evaluate RheOFormer's ability to model Oldroyd-B and Giesekus fluids in 4:1 contraction flow, a benchmark flow geometry for the assessment of computational fluid dynamics models in solving viscoelastic flows. 

For the Oldroyd-B case, the dataset was generated via numerical simulations and consisted of velocity components ($u_x, u_y$) and stress components ($\sigma_{xx}, \sigma_{yy}, \sigma_{xy}$) at $5425$ spatial locations, spanning time steps from $t = 0$ to $5\,\text{s}$, with a temporal resolution of $\Delta t = 0.2\,\text{s}$. The inlet velocity, used as the varying input condition, ranges from $0.01$ to $2.0$ $\si{\centi\meter\per\second}$ across 64 training samples, uniformly distributed across this range. 
Given a fixed relaxation time of $\lambda = 0.1\,\text{s}$, this corresponds to Weissenberg numbers ranging from $Wi=0.004-0.8$. 
Each training sample corresponds to a single simulation at a fixed $Wi$, and test samples are evaluated at uniformly spaced, held-out $Wi$ values within the same range.

\begin{figure*}
\centering
\includegraphics[width=\textwidth]{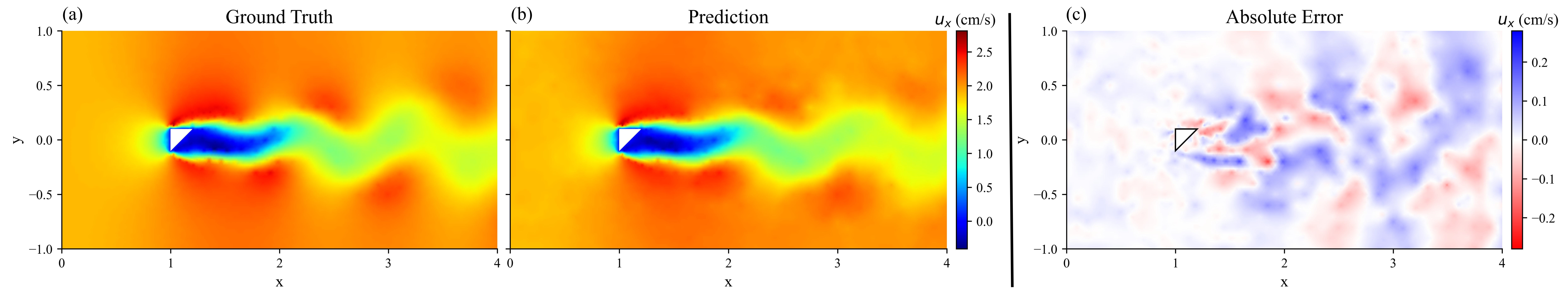}
\caption{RheOFormer prediction of an Oldroyd-B fluid's wake dynamics behind a triangular obstacle. (a) Ground truth velocity, $u_x$, map at $Wi=0.94$ and $Re = 94$, (b) RheOFormer predictions at the same $Wi$ and $Re$, 
and (c) absolute error field between the ground truth and predicted $u_x$ at the last time step. The model receives only the first five temporal snapshots as input and predicts the remaining time steps. The absolute error remains small throughout the domain, with the largest discrepancies localized near the wake region and shear layers downstream of the obstacle, where large velocity gradients are present.}
\label{fig:Triangle}
\end{figure*}

In order to assess the limits of RheOFormer's ability in predicting nonlinearities observed in contraction flows, the Giesekus fluid was intentionally made more challenging by increasing the relaxation time to $\lambda = 1\,\text{s}$, resulting in $Wi=0.1-4.2$ and using only 24 uniformly spaced training samples. For both fluids however, RheOFormer was trained on all available physical fields (velocities and stress components) to model the full set of coupled dynamics. During inference, the model received only the first ten time snapshots (up to $t = 1.8\,\text{s}$) and was tasked with predicting the remaining temporal evolution of the flow. Internally, the encoder extracted spatio-temporal patterns from the inputs using self-attention mechanisms, layer normalization, and feed-forward neural networks. The decoder, in turn, received the encoded representation along with the coordinates of the desired output points and the number of future time steps to predict. It employed cross-attention to relate encoded features to output targets and marched forward in time step-by-step to reconstruct the full solution trajectory.

Figure \ref{fig:channel} presents the predicted and reference velocity magnitude fields $|\mathbf{u}|$ for both fluids. 
The displayed cases ($Wi = 0.35$ for the Oldroyd-B fluid and $Wi = 3.9$ for the Giesekus fluid) were not included in the training set.
In each subfigure, the upper half displays the numerical ground truth, while the lower half shows RheOFormer predictions. For both the Oldroyd-B fluid (Figure \ref{fig:channel}a), and the Giesekus fluid (Figure \ref{fig:channel}b), RheOFormer accurately simulates the entire flow, evident from symmetrical flow structures and the downstream velocity profiles. Supplementary Movie~S1 shows the velocity magnitude, shear stress, and normal stress components for the Giesekus fluid with the same flow conditions as shown in Figure \ref{fig:channel} for $Wi=3.9$.As clearly evident in Supplementary Movie~S1, RheOFormer consistently predicts the flow velocities and the underlying stress (shear and normal) profiles for the contraction flow for the entire time of simulation.

Having benchmarked RheOFormer as an accurate viscoelastic solver, next we employ the architecture to model more complex flow geometries involving the wake formation behind a triangular obstacle. The dataset was generated numerically including $u_x, u_y, \sigma_{xx}, \sigma_{yy}, \sigma_{xy}$ for an Oldroyd-B fluid over a time of $t = 0\text{--}10~\mathrm{s}$, with $\Delta t = 0.2~\mathrm{s}$. The $Wi$ number was varied from 0.1 to 1.0 across 200 uniformly spaced training samples. All provided physical variables (velocities and stress tensors) are included in the training step to effectively learn underlying physical dependencies and produce highly accurate predictions. The model received only the first five temporal snapshots as input and was asked to predict the remaining 46 time steps. Figure \ref{fig:Triangle} shows the predicted and ground truth $u_x$ fields, along with the corresponding absolute error at $Wi = 0.94$, a value not included in the training set.
The model successfully reproduces key flow features such as the formation of wake regions, shear layers, and vortex structures downstream from the obstacle. Visually, the predicted flow patterns closely reproduce key physical features observed in the ground truth data, such as the wake formation and characteristic shear-layer structures well past the triangular obstacle.

To quantitatively evaluate the prediction accuracy, we examine the pointwise error field $(u_x^{\mathrm{GT}} - u_x^{\mathrm{pred}})$ shown in Figure~\ref{fig:Triangle}c. The error magnitude remains below $0.2$~\si{\centi\meter\per\second} throughout the domain, corresponding to less than $7\%$ of the peak streamwise velocity ($u_{x,\mathrm{max}} \approx 3$~\si{\centi\meter\per\second}). The largest discrepancies are localized near the wake region and shear layers immediately downstream of the obstacle, where steep velocity gradients are present and small spatial shifts between predicted and reference fields produce locally elevated errors. Away from these regions, the error is negligibly small.

While relative error is generally the preferred metric for quantifying prediction accuracy, a pointwise relative measure $\left|\frac{u_x^{\mathrm{GT}} - u_x^{\mathrm{pred}}}{u_x^{\mathrm{GT}}}\right|$ becomes ill-conditioned in regions where the ground truth velocity approaches zero, such as near the obstacle walls and in the far wake, producing artificially inflated values that do not reflect meaningful prediction failures. The pointwise error map in Figure~\ref{fig:Triangle}c confirms that these regions carry negligible physical discrepancy, motivating the use of the absolute metric here.

Supplementary Movie~S2 and Movie~S3 show the time evolution of the velocity components and the shear stress field for the case shown in Figure~\ref{fig:Triangle}, confirming that the wake structures are reproduced over the entire prediction horizon. However, small deviations are visible in the full temporal evolution of $\sigma_{xy}$. The predicted shear stress field exhibits fine scale variation in the low-magnitude regions away from the wake, where the ground truth is smooth and close to zero. In absolute terms these deviations are of order $10^{-5}$~Pa, three to four orders of magnitude below the peak stress magnitude, and are therefore physically negligible while the logarithmic color scale amplifies them visually. The amplitude of this variation does not grow over the prediction horizon (consistent with Figure~S7), confirming that it reflects the resolution limit of the predicted field in regions of vanishing stress rather than any instability of the latent propagator. Additionally, Supplementary Figure~S1 shows the velocity maps over a wide range of applied $Wi$, benchmarked against the ground truth simulations. These results demonstrate RheOFormer's ability to accurately simulate complex fluids in complex flow geometries.

\subsection{Data Efficiency and Generalization} 
Returning to the rheometric flow benchmarks introduced above, we now investigate RheOFormer's sensitivity to training set size and its ability to extrapolate beyond the training distribution. Two complementary studies were conducted for both the TEVP and Giesekus constitutive models: (i)~the number of training samples was progressively varied while holding all other hyperparameters fixed, and (ii)~models trained with a GRF output scale of $\mathcal{S} = 1.0$ were evaluated on samples drawn from distributions with larger output scales, effectively exposing the model to deformation rates larger than those encountered during training.

\begin{figure}[t]
    \centering
    \includegraphics[width=8.7cm]{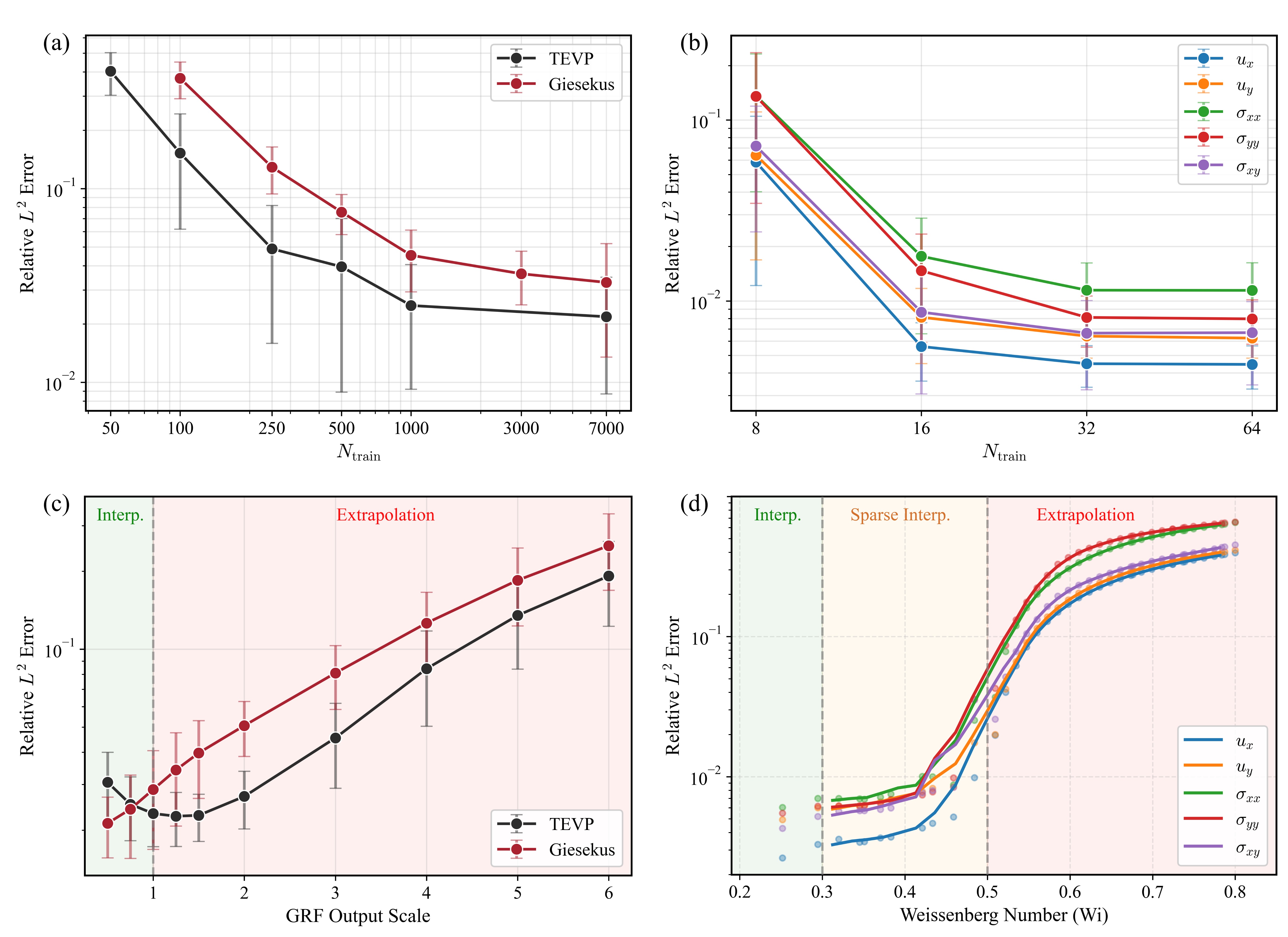}
    \caption{
    Data efficiency and extrapolation performance of RheOFormer.
    (a)~Relative $L^2$ error as a function of training set size $N_\mathrm{train}$ for the TEVP and Giesekus models, evaluated on simple shear tests not seen during training. 
    (b)~Relative $L^2$ error per field component as a function of $N_\mathrm{train}$ for the Oldroyd-B 4:1 contraction flow, evaluated on held-out $Wi$ values within the training range.
    (c)~Relative $L^2$ error as a function of GRF output scale $\mathcal{S}$ for the TEVP and Giesekus models. Both models were trained at $\mathcal{S}=1.0$ (dashed vertical line); green and red shading indicate the interpolation and extrapolation regimes, respectively. 
    (d)~Relative $L^2$ error per field component versus Weissenberg number for the Oldroyd-B contraction flow. The model was trained on 32 samples spanning $Wi\in[0.004,\,0.3]$ (green-shaded region). To probe generalization under sparse data coverage, 10 additional training samples were included in the range $Wi\in[0.3,\,0.5]$ (yellow-shaded region), where the model is tested on held-out $Wi$. 
    The red-shaded region denotes extrapolation beyond the training boundary ($Wi>0.5$). 
    All errors are mean relative $L^2$ errors computed over the full spatial and temporal extent of the predicted trajectory.
    Error bars denote one standard deviation across test samples.}
    \label{fig:ntrain}
\end{figure}

For the TEVP model, RheOFormer was trained on $N_{\mathrm{train}} = 50$ to $7000$ GRF-generated shear rate profiles and evaluated on an oscillatory shear test unseen during training. As shown in Figure~\ref{fig:ntrain}a, the relative $L^2$ error decreases by nearly an order of magnitude from $N_{\mathrm{train}} = 50$ to $250$ and reaches near-saturation at $N_{\mathrm{train}} = 500$, beyond which additional data yields diminishing returns.  A per-channel analysis (Figure~S2) reveals that the shear stress $\sigma_{12}(t)$ consistently exhibits higher relative error than the structure parameter $\lambda(t)$ across all training set sizes and test protocols, reflecting the stronger nonlinearity of the stress response. Representative time series (Figure~S3) confirm that at $N_{\mathrm{train}} = 50$ the errors concentrate in the amplitude and phase of $\sigma_{12}(t)$; by $N_{\mathrm{train}} = 500$ both channels are nearly indistinguishable from the exact solution.

The tensorial Giesekus model, with four input velocity gradient components and three coupled stress outputs, presents a more demanding learning task. The model was trained on $N_\mathrm{train}=100$ to 7000 GRF-generated random samples and evaluated on simple shear flow. As shown in Figure~\ref{fig:ntrain}a, the Giesekus model requires substantially more data than the TEVP case to achieve comparable accuracy. At $N_{\mathrm{train}} = 100$ the relative $L^2$ error exceeds $0.3$, and the time series predictions exhibit high-frequency oscillatory artifacts (Figure~S5). The error drops below $0.05$ at approximately $N_{\mathrm{train}} = 1000$ and plateaus by $N_{\mathrm{train}} = 3000$, with predictions at $N_{\mathrm{train}} = 7000$ nearly indistinguishable from the exact solution. A per-channel breakdown (Figure~S4) reveals that the normal stress components $\sigma_{11}$ and $\sigma_{22}$ are consistently harder to predict than $\sigma_{12}$, reflecting the strong coupling introduced by the quadratic Giesekus nonlinearity ($\alpha\,\boldsymbol{\sigma} \cdot \boldsymbol{\sigma} / G_0$). Overall, RheOFormer learns scalar TEVP constitutive behavior from as few as ${\sim}250$ samples, while the tensorial Giesekus model requires ${\sim}3000$ samples, which is still efficient given the dimensionality of the input and output spaces involved.

To investigate whether these data efficiency trends extend to full spatio-temporal PDE problems, an analogous study was conducted for the Oldroyd-B contraction flow. Figure~\ref{fig:ntrain}b shows the relative $L^2$ error per field component as a function of training set size, where $N_\mathrm{train}$ was varied from 8 to 64 while holding all other settings fixed. With only 8 training samples, velocity components exhibit relative errors of approximately $0.05$, while the stress components reach $0.15$. A sharp improvement is observed from $N_\mathrm{train}=8$ to 16, with the relative error of most fields dropping below $0.02$. Beyond $N_\mathrm{train}=32$, the gains are marginal, indicating that the model has effectively saturated in accuracy for this geometry. 

Across all training set sizes, $\sigma_{xx}$ consistently exhibits the highest error among all field components, reflecting the steep normal stress gradients that develop near the contraction corners. A similar convergence trend is observed for the flow past a triangular obstacle (Figure~S6), where errors plateau by approximately $N_\mathrm{train}=50$. The temporal evolution of errors (Figure~S7) further shows that with sufficient training data ($N_{\mathrm{train}} \geq 50$) the propagator remains stable over the full 46-step prediction horizon, whereas with insufficient data the predictions progressively diverge from the ground truth as the model propagates further in time.

To assess generalization beyond the training distribution, models trained on GRF-generated samples with a fixed output scale of $\mathcal{S} = 1.0$ were evaluated on test data generated with output scales ranging from $\mathcal{S} = 0.5$ to $6.0$, where larger $\mathcal{S}$ produces input profiles with proportionally larger deformation rate amplitudes. 

For the TEVP model (Figure~\ref{fig:ntrain}c), the relative $L^2$ error remains nearly flat from $\mathcal{S} = 0.5$ to $1.5$, stays below $0.04$ up to $\mathcal{S} = 2.0$, and below $0.1$ up to $\mathcal{S} = 4.0$. Even at $\mathcal{S} = 6.0$, where deformation rates are six times larger than those seen during training, the model retains qualitatively correct predictions. Time series comparisons (Figure~S9) confirm that up to $\mathcal{S} = 6.0$ errors appear as modest amplitude offsets rather than qualitative misrepresentation of the dynamics, though predictive reliability breaks down by $\mathcal{S} = 10.0$. The growing error bars at large scales reflect the increased variability in deformation rate magnitudes across individual GRF realizations: at higher output scales, some test samples contain peak rates far outside the training range while others remain moderate, leading to a wider spread in prediction accuracy.

The Giesekus model exhibits a similar profile (Figure~\ref{fig:ntrain}c): the relative $L^2$ error remains below $0.03$ within the training distribution, below $0.07$ up to $\mathcal{S} = 2.0$, and below $0.15$ up to $\mathcal{S} = 4.0$. Notably, all three stress tensor components degrade at comparable rates (Figure~S10), in contrast to the training set size study where $\sigma_{11}$ was disproportionately difficult. This suggests that with sufficient training data, the model learns the tensorial coupling structure well, and that extrapolation errors arise from encountering stress magnitudes outside the training range rather than from a failure to capture inter-component dependencies. At $\mathcal{S} = 6.0$, visible deviations appear as underestimation of peak stress amplitudes, but the temporal structure is preserved (Figure~S11), though predictive reliability breaks down by $\mathcal{S} = 10.0$.

For the canonical flow case, extrapolation in Wi space was probed by training the Oldroyd-B contraction model on only the 32 samples falling within $Wi\in[0.004,\,0.3]$ plus 10 additional samples in the range $Wi\in[0.3,\,0.5]$.
Figure~\ref{fig:ntrain}d reports the relative $L^2$ error on held-out samples as a function of $Wi$ across three regimes. Within the original training range ($Wi<0.3$), all field components maintain errors below 0.005. In the sparse interpolation regime ($Wi\in[0.3,\,0.5]$), despite having only 10 training samples, errors remain between 0.005 and 0.05, indicating that RheOFormer maintains high prediction accuracy even with limited coverage of the parameter space. Near the upper boundary of this range ($Wi\approx 0.45$--$0.5$), a sharp increase in error is observed as the model transitions into the extrapolation regime. Beyond the training boundary ($Wi>0.5$), errors increase monotonically, reaching $30\%-60\%$ by $Wi=0.8$. 
The velocity component $u_x$ degrades more gradually than the stress fields, consistent with the smoother spatial structure of velocity relative to the stress components in viscoelastic contraction flows. For the flow past a triangular obstacle, extrapolation performance degrades more rapidly due to the stronger nonlinearities associated with wake formation and unsteady shear layer dynamics at elevated $Wi$ (presented in Figure~S12).

\section{CONCLUSION}\label{sec12}

In this study, we presented RheOFormer, an attention-based neural operator surrogate for non-Newtonian fluid dynamics, and systematically characterized its accuracy, data requirements, and failure modes. Through detailed benchmarking against ground truth (numerical) solutions, our transformer-based framework demonstrated high levels of accuracy in predicting/modeling a wide range of complex fluids in rheometric flows as well as two-dimensional flow geometries. Namely, viscoelastic (Giesekus and Oldroyd-B fluids) and thixotropic elasto-viscoplastic fluids were modeled in various flow kinematics. By effectively integrating self-attention, cross-attention, and latent time-marching mechanisms, RheOFormer showed remarkable efficiency in capturing both scalar and tensorial stresses in complex fluids exposed to different flowing conditions. We showed that the architecture can learn rich operator mappings from limited data, with accuracy saturating near $N_\mathrm{train} \approx 250$ samples for the scalar TEVP response and $N_\mathrm{train} = 32$ simulations for the
Oldroyd-B contraction flow. The same architecture maintains this accuracy across distinct constitutive models, geometries, and flow protocols without any modification, emphasizing its flexibility and robustness across varied physical contexts and geometric complexities. Furthermore, the latent-space propagation strategy substantially reduces memory cost relative to fully unrolled training while preserving prediction accuracy and long-time stability.

Equally informative are the boundaries identified here. Within the sampled Weissenberg range, the contraction model maintains mean relative $L^2$ errors below [0.01] across all field components, and degrades only modestly under sparse parameter coverage, remaining between 0.005 and 0.05 with as few as ten training samples in the sparsely sampled interval. Beyond the training boundary, errors grow monotonically to [30--60]\%\ and predictions become unreliable. Stress components consistently degrade before velocity fields, reflecting their sharper spatial gradients and wider dynamic range. For the rheometric cases, predictions remain qualitatively correct up to six times the training deformation rate amplitude but break down after that.

These findings position RheOFormer as a robust and efficient tool for surrogate modeling in a broad range of applications. It should be noted that in the current framework, each geometry is treated with its own trained model, a characteristic shared by most neural operator approaches applied to spatially complex domains. While the model receives spatial coordinates and boundary indicator variables as inputs, which help capture local features near solid boundaries, these are specific to a given domain and do not constitute a transferable geometry representation. Nonetheless, extending the framework to generalize across geometries would likely require richer geometry encodings capable of representing diverse domain topologies within a unified input space. This remains an open challenge for the broader neural operator community.
Despite the limitations, given the pace of developments in generative AI methodologies, this approach presents a practical pathway to democratization of highly technical and detailed non-Newtonian fluid dynamics in any and all processes involving soft materials and flow.

\section*{Supplementary Material}

See the supplementary material for details of the CFD setups used to generate the reference data for both flow geometries, additional accuracy and generalization studies, and three movies comparing the temporal evolution of the predicted velocity and stress fields against the ground truth simulations.

\begin{acknowledgments}
M.S. and S.J. would like to acknowledge the support of DoD MURI Award N00014-23-1-2499.

\end{acknowledgments}

\section*{Data Availability Statement} 
The code for RheOFormer, including hyperparameter configurations, the simulated datasets, training scripts and evaluation pipelines, is publicly available in the GitHub repository at \url{https://github.com/procf/RheOFormer}.

\appendix*

\section{Constitutive Models}\label{sec:CM}

\subsection{Thixotropic Elasto-Viscoplastic (TEVP) Model}\label{TEVP} 
The TEVP model characterizes the temporal evolution of shear stress within structured materials through two coupled ordinary differential equations (ODEs)\cite{larson2015constitutive, jamali2022mnemosyne}. The first equation relates the internal shear stress to the deformation rate through:

\begin{equation}
\begin{split}
\dot{\sigma}_{12}(t)
= \frac{G}{\eta_s + \eta_p}\Big[
-\sigma_{12}(t) + \sigma_y \lambda(t) \\
+ \left[\eta_s + \eta_p \lambda(t)\right] \dot{\gamma}_{12}(t)
\Big]
\end{split}
\end{equation}

where $G$ denotes the elastic modulus (\si{\pascal}), $\sigma_{y}$ is the yield stress (\si{\pascal}), $\eta_s$ and $\eta_p$ are the solvent (background) and plastic viscosities (\si{\pascal}$\cdot$\si{\second}), respectively, $\dot{\gamma}_{12}(t)$ is the shear rate ($s^{-1}$), and $\lambda(t)$ represents the time- and rate-dependent dimensionless structure parameter. The structure parameter $\lambda(t)$ quantifies the instantaneous degree of microstructure within the fluid under shear flow, distinguishing fully fluidized ($\lambda = 0$) and fully structured states ($\lambda = 1$)\cite{de2011thixotropic}. The temporal evolution of the structure parameter can be written as:

\begin{equation}\label{eq:6}
\dot{\lambda}(t) = k_{+}(1 - \lambda(t)) - k_{-} \lambda(t) \lvert\dot{\gamma}_{12}(t)\rvert
\end{equation}
In this expression, $k_+$ ($s^{-1}$) and $k_-$ ($s^{-1}$) denote the structural buildup rate under quiescent conditions and structural breakdown rate under shear flow, respectively, and $\lvert\dot{\gamma}_{12}(t)\rvert$ represents the absolute shear rate. Consequently, $\lambda(t)$ evolves based on competition between structure breakdown due to shear flow and structure formation arising from the intrinsic tendency of fluid components to aggregate. Equations above merely represent typical TEVP mechanics, and many other forms of both the stress-strain coupling and the structural evolution can be realized\cite{larson2019review}. It should also be noted that while in this work we have only focused on the shear component of the stress tensor, TEVP constitutive models are fully generalizable to tensorial descriptions and can be solved for normal stresses as well. In this work, shear rate $\dot{\gamma}_{12}(t)$ is randomly varied during training to demonstrate the feasibility of Neural Operators in learning a broad family of constitutive model behaviors. These results can be easily generalized to incorporate more complex constitutive models.

\subsection{Giesekus Model} \label{Giesekus}
The tensorial two-dimensional Giesekus model is a common choice for viscoelastic fluids, as it represents the nonlinear viscoelastic/memory effects through upper-convected derivative function. By decoupling the solvent and polymer stresses, the model includes a mobility factor that captures the nonlinear dynamics at large stresses or strains\cite{giesekus1982simple}. The Giesekus model is expressed in its general form as:
\begin{equation}  
\boldsymbol{\sigma} + \tau_1 \overset{\nabla}{\boldsymbol{\sigma}} + \frac{\alpha}{G_0}\boldsymbol{\sigma}\cdot\boldsymbol{\sigma} = G_0\tau_1\left(\dot{\boldsymbol{\gamma}} + \tau_2\overset{\nabla}{\dot{\boldsymbol{\gamma}}}\right)
\end{equation}

In the above equation, $\boldsymbol{\sigma}$ and $\dot{\boldsymbol{\gamma}}$ denote the stress and deformation-rate tensors, respectively, while $\nabla$ represents the upper-convected derivative\footnote{For a detailed discussion on the advantages of convected coordinate systems in achieving frame-invariant expressions of time-varying stresses and deformations, see\cite{morrison2001understanding}}. Model parameters include relaxation time $\tau_1 (s)$, retardation time $\tau_2 (s)$, the elastic modulus $G_0 (\si{\pascal})$, and the mobility factor $\alpha$. The parameters $\tau_1$ and $\tau_2$ critically determine the transient response duration of the fluid under flow\cite{vlassopoulos1995generalized}. The mobility factor $\alpha$ allows the model to effectively describe shear thinning and elongational thickening behaviors, prevalent in polymeric systems, and which simpler constitutive models such as the Oldroyd-B model do not adequately capture.

The deformation rate tensor is derived from the velocity gradient tensor $\nabla \mathbf{u}$ as:
\begin{equation}{
\dot{\gamma}_{ij} = \nabla u_{ij} + \nabla u_{ji}
}\end{equation}
where $\nabla u_{ij}$ denotes the components of the velocity gradient tensor. In the rheometric tests presented in this work, the four independent velocity gradient components ($\nabla u_{11}$, $\nabla u_{22}$, $\nabla u_{12}$, $\nabla u_{21}$) serve as inputs to RheOFormer, and the stress tensor components ($\sigma_{11}$, $\sigma_{22}$, $\sigma_{12}$) constitute the predicted outputs.

\subsection{Oldroyd-B Model} \label{Oldroyd}
The Oldroyd-B constitutive model provides another foundational model for describing the viscoelastic behavior of polymeric fluids. Unlike the Giesekus model, the Oldroyd-B formulation does not incorporate a mobility factor and thus cannot explicitly capture anisotropic drag effects\cite{oldroyd1950formulation}. Its tensorial formulation is typically represented as:

\begin{equation}
\boldsymbol{\sigma} + \tau_1 \overset{\nabla}{\boldsymbol{\sigma}} = G_0\tau_1\left(\dot{\boldsymbol{\gamma}} + \tau_2\overset{\nabla}{\dot{\boldsymbol{\gamma}}}\right)
\end{equation}

Here, similar to the Giesekus model, $\boldsymbol{\sigma}$ denotes the stress tensor, $\dot{\boldsymbol{\gamma}}$ the deformation-rate tensor, and $\nabla$ is the upper-convected derivative. The Oldroyd-B model is characterized by two essential time constants: the relaxation time $\tau_1$ and the retardation time $\tau_2$, which together define the fluid's response dynamics to deformation\cite{oldroyd1958non}. While effectively capturing linear viscoelastic behaviors such as stress relaxation and creep recovery, the Oldroyd-B model is limited in describing nonlinear phenomena like shear-thinning or strain hardening that the more advanced Giesekus model addresses through the introduction of the mobility factor $\alpha$. Nevertheless, the Oldroyd-B model remains an important benchmark due to its analytical tractability and widespread application in characterizing dilute polymeric solutions under simpler flow conditions.

\bibliography{aipsamp}

\end{document}